# CMAWRNet: Multiple Adverse Weather Removal via a Unified Quaternion Neural Architecture

Vladimir Frants, Sos Agaian, Fellow, IEEE, and Karen Panetta, Fellow, IEEE, Member, IEEE, Peter Huang

*Abstract*—**Images used in real-world applications such as image or video retrieval, outdoor surveillance, and autonomous driving suffer from poor weather conditions. When designing robust computer vision systems, removing adverse weather such as haze, rain, and snow is a significant problem. Recently, deep-learning methods offered a solution for a single type of degradation. Current state-of-the-art universal methods struggle with combinations of degradations, such as haze and rain-streak. Few algorithms have been developed that perform well when presented with images containing multiple adverse weather conditions. This work focuses on developing an efficient solution for multiple adverse weather removal using a unified quaternion neural architecture called CMAWRNet. It is based on a novel texture-structure decomposition block, a novel lightweight encoder-decoder quaternion transformer architecture, and an attentive fusion block with low-light correction. We also introduce a quaternion similarity loss function to preserve color information better. The quantitative and qualitative evaluation of the current state-of-the-art benchmarking datasets and real-world images shows the performance advantages of the proposed CMAWRNet compared to other state-of-the-art weather removal approaches dealing with multiple weather artifacts. Extensive computer simulations validate that CMAWRNet improves the performance of downstream applications such as object detection. This is the first time the decomposition approach has been applied to the universal weather removal task.**

## I. INTRODUCTION

INTELLIGENT Transportation Systems (ITS) encompass many technologies to enhance transportation safety, efficiency, and reliability. These technologies range from advanced driver-assistance systems (ADAS) and autonomous vehicles to transport surveillance and traffic management systems.

Statistics reveal the profound impact of traffic accidents worldwide, with an alarming annual death toll of approximately 1.3 million people and over five million injuries in the United States alone. This underscores the critical need for effective ITS solutions. Weather-related incidents, responsible for about 16% of all vehicular fatalities in the US, highlight the urgency of addressing challenges posed by adverse weather conditions like haze, rain, and snow. Images captured under these conditions typically accompany low-lighting conditions and adherent raindrops. To improve the situation, adverse weather image restoration has been extensively studied in the forms of dehazing [1]-[5], deraining [6]-[10], snow removal [11]-[15], etc. These weather conditions significantly reduce the visibility of details in images, negatively affecting the performance of computer vision algorithms, including object detection, semantic segmentation, and anomaly detection.

Prior-based methods for weather removal emerged [1]-[2], [6]-[7], [11]-[12] focused on a single weather condition. However, it is required to address multiple types of weather simultaneously for real-world scenarios. These additional requirements complicate the design and increase the computational requirements of real-world vision systems for video surveillance and autonomous robotics. Each weather condition requires distinct prior, based on certain assumptions, and when the assumptions are not satisfied, the performance degrades. Recently, deep-learning-based methods have become more popular due to their inherited ability to learn the priors from the data. There are trade-offs and pros and cons to using any of these approaches.

On the one hand, specialized methods demonstrate excellent performance on synthetic datasets, but the generalization ability depends on the quality and the size of the dataset, as adverse weather removal is an ill-posed problem. For many problems, such as rain streaks and snow removal, collecting large, high-quality datasets is impossible, which leads to poor performance on real-world images [25], [27].

Recent methods, including MPR-Net [16], HINet [17], Swin-IR [18], U-former [19], and Restormer [20], are designed with general image restoration in mind. These methods are validated on multiple tasks, including adverse weather conditions. Nevertheless, a separately trained model is used for each condition, and removing the multiple degradations is impossible with a single set of weights.

The literature on universal methods capable of removing multiple weather conditions in one step is very limited. The first work in this direction is Li et al. All-in-One Bad Weather Removal Network [21]. It proposes an end-to-end trained CNN with multiple convolutional encoders, one for each condition: snow, raindrops, and a combination of rain-streak and haze. Despite the ability to handle various degradations, All-in-One can only tackle one degradation at a time and has a large number of parameters due to multiple encoders. TransWeather employs a similar technique but uses a single visual transformer encoder and trainable weather-type query embedding to handle various degradations [22]. The network is effective in processing one weather degradation at a time. Still, performance significantly drops in the case of multiple degradations, such as severe haze and rain streaks. Chen et al. propose multiple adverse weather removal methods trained with the help of transfer learning [23]. For each weather condition, a separate large teacher network is trained. Then, the knowledge of several teacher networks is transferred to a more compact student network. These methods indicate significant progress in universal weather removal but are complex and computationally demanding. More information on current progress in weather removal could be found in surveys [24]-[28]. Below, we summarize the main limitations of the current state-of-the-art weather removal algorithms:

1. *Over-smoothing, unnatural color, inability to handle low-light images.* Over-smoothing is common in textured background regions due to its complex nature and high variability, such as degradations caused by snow, rain streaks, and raindrops. The presence of the haze generally distorts the color information and further contributes to the over-smoothing of the restored image. Even though color plays an important role, deep-learning-based methods are usually trained with objective functions that do not consider the properties of color spaces and relation among color channels or completely ignore the color. Moreover, the standard way to evaluate the methods is the application of pixel-wise metrics SSIM and PSNR on the Y channel of the YCrCb color space, which does not consider the color information [46].

2. *The limited ability to model complex patterns, such as multiple overlapping rain streaks or snowflakes and combinations of the haze and rain streaks.* Raindrops and snow accumulation over a distance lead to fog effects, introducing complex visual artifacts that require customized priors for effective restoration. This variability poses a significant challenge for universal weather removal models.

3. *Inability to effectively process multiple degradations.* Despite the recent progress, even universal methods cannot effectively remove several degradations presented in the same image.

In this work, we propose a universal weather removal method, CMAWRNet, incorporating quaternion color representation, separate texture and structure content processing, and attentive fusion with low-light correction to address the image details visibility degradation in adverse weather and lighting conditions. Quaternions are four-dimensional extensions of complex numbers. A quaternion has one real and three separate imaginary components and enables the processing of color value as a single entity [30]. The quaternion color representation replaces a triplet of color channels (R, G, B) with a single quaternion number, preserving relationships between R, G, and B [21], [22], [30]. Quaternion neural networks take a quaternion-valued image as an input and produce quaternion-valued output using rules of quaternion algebra. Studies have shown that quaternion neural networks (QNNs) (i) are effective in cases when the real-valued neural networks (RNNs) fail to capture the color information [36], [47], [48], (ii) have already shown certain advantages over RNNs in speech, image compression, objective image quality assessment, and image classification [30]-[36], (iii) deliver state-of-the-art performance on various tasks by reducing the number of training parameters and explicitly modeling the interchannel correlation [29], and (iv) enable effective learning of the interchannel and spatial relations between multiple input feature maps. So, a QNN makes a reasonable basis for a unified architecture for multiple adverse weather removal, as presented in this paper.

To effectively handle multiple adverse weather conditions, it's important to recognize how different phenomena affect specific parts of an image. Inspired by Retinex theory, we decompose the input image into illumination ($L$) and reflectance ($R$) components, where $U = R \circ L$ and $\circ$ denotes element-wise multiplication. The illumination component captures the smooth, overall scene structure, whereas the reflectance component contains the fine textures and details of objects. By processing these components separately, we can target specific degradations more effectively. Haze primarily affects the illumination component by reducing visibility and contrast across the scene. In contrast, rain streaks, raindrops, and snowflakes mainly corrupt the reflectance component by introducing high-frequency noise that obscures textures. Leveraging local derivatives—where larger values indicate texture changes and smaller ones correspond to smooth structures—allows for accurate decomposition of the image content. This approach enables us to apply specialized enhancement techniques to each component, resulting in improved restoration quality.

This paper aims to develop a quaternion convolutional neural network called CMAWRNet for multiple degradation removal to tackle the abovementioned challenges. Our main contributions are:

1. We introduce:
- *CMAWRNet*: a computationally efficient unified quaternion network architecture designed to remove multiple degradations caused by adverse weather conditions.
- A novel quaternion similarity loss function that preserves color information while optimizing the network.
- DNet: A sub-network for image decomposition into texture and structure components. TNet: A lightweight transformer decoder-encoder sub-network that leverages the unique properties of quaternion neural networks. FNet: An attention-based quaternion neural network (QNN) that fuses texture and structure information while simultaneously correcting low-light conditions.

2. We provide comprehensive experimental results demonstrating the effectiveness of CMAWRNet in removing multiple degradations from input images. These experiments were conducted on multiple datasets and real-world images, showing superior performance in texture detail preservation and overall image quality compared to state-of-the-art methods.

The remainder of the paper is organized as follows. Section II presents an overview of previous work on weather removal. Section III describes the proposed CMAWRNet architecture. Section IV presents and discusses the experimental results. Section VI concludes the paper.

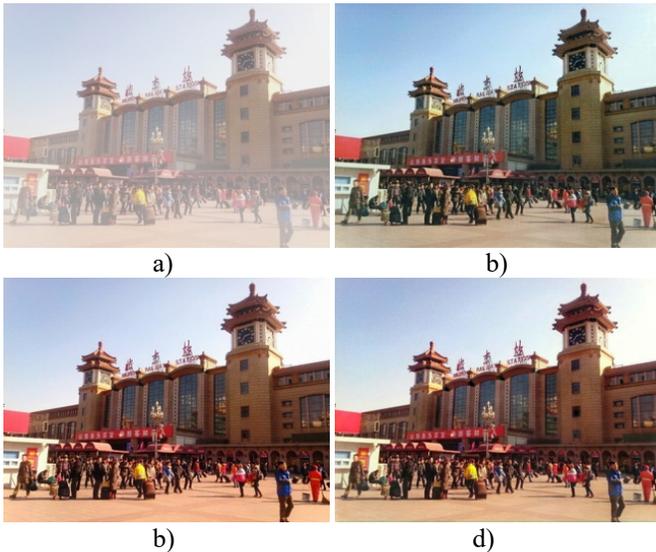

**Fig. 1.** Comparison between RNN and QNN for image restoration. a) Input image with degradation. b) Output from the RNN (0.3M parameters). c) Ground truth. d) Output from the QNN (0.078M parameters) shows improved color fidelity and reduced geometric artifacts while significantly lowering the number of parameters.

## II. PREVIOUS WORK

### A. Adverse Weather Removal

Existing image restoration approaches include single weather removal algorithms for dehazing [1]-[5], deraining [6]-[10], desnowing [11]-[15], adherent raindrop removal [9], [53], multi-degradation removal [16], [60], and universal methods (all in one strategy, addressing multiple degradations at the same time) [22], [23].

**Single weather conditions removal** methods focus on addressing a single type of adverse weather condition, such as rain or fog. *For haze removal,* Cai et al. designed CNN for medium transmission map estimation and removed the haze using the atmospheric scattering model [3]. Li et al. developed a lightweight CNN for end-to-end haze removal without explicitly estimating transmission function [4]. Mei et al. propose a more sophisticated end-to-end dehazing architecture employing UNet-like architecture and progressive feature fusion [39]. FFA-Net introduces an attentive feature fusion to give more weight to essential features and improve the dehazing result [43]. MSNet uses multiscale feature maps for higher spatial resolution and better contrast [45]. RefineDNet employs a two-stage strategy using the dark channel prior for visibility restoration and weakly supervised CNN to remove artifacts introduced by the dehazing procedure and to improve the realness measure [5]. *For rain removal,* Fu et al. introduce a dual graph convolutional network with a long-range contextual information aggregation mechanism to process long rain streaks efficiently [50]. Wang et al. narrow the domain gap between the real-world rain images and synthetic ones used for training by introducing novel physics-based rain generation procedures [51]. Qian et al. introduced a dataset for adherent raindrop removal and an attentive GAN for a single image raindrop removal [9]. Quan et al. introduce CNN with a double attention mechanism for the accurate localization of the raindrops and channel re-calibration to improve the processing of raindrops of various shapes [52]. Quan et al. propose CCN – a complementary cascaded network architecture, to remove rain streaks and raindrops in a complementary fashion via neural architecture search [53]. *For snow removal,* Liu et al. proposed a synthetic dataset Snow100K, and DesnowNet – a multistage, multiscale CNN for opaque and translucent snow particles removal [13]. Chen et al. introduce a novel snow model, including the veiling effect and a transparency-aware convolutional architecture JSTASR [54]. DDMSNET uses semantic and geometry information in a three-stage coarse-fine snow removal framework [61]. Ye et al. developed an efficient pyramid network for real-time high-resolution image snow removal [14]. Chen et al. propose a scale-aware transformer encoder-decoder network with context interaction [15]. Despite the excellent performance of the single weather conditions removal methods, they often include weather-specific blocks and priors and generally do not perform well on other tasks.

**Multi-degradation removal** offers a general image enhancement architecture that can be repurposed for any specific degradation. MPRNet uses a three-stage framework with shared features [16]. Chen et al. investigate the role of the normalization layer in the performance of multiscale, multistage architecture on low-level image processing tasks [17]. Swin-IR offers a baseline transformer-based architecture instead of commonly used CNNs for image super-resolution, JPEG-compression artifact reduction, low-light image enhancement, etc. [18]. Wang et al. introduce a novel locally-enhanced window transformer block and a learnable multiscale restoration modulator in an architecture called U-former [19]. Zamir et al. introduce Restormer – an efficient and effective transformer-based architecture for low-level image processing [20]. These methods deliver better or comparable performance to weather-specific methods, but each specific task requires a distinct set of weights and, sometimes, a specialized training procedure.

**Universal (All-in-One) methods** handle multiple weather conditions employing fixed architecture and weights. As a first attempt to develop a universal multi-weather removal network, Li et al. proposed the All-in-One method [21]. All-in-one takes an image degraded by any weather condition and predicts a clean image. A separate encoder, determined with a neural architecture search, is used for each weather type. TransWeather builds up on the same idea, but instead of multiple convolutional encoders, a single visual transformer encoder and a decoder with weather-type embedding are applied [22]. Chen et al. introduce a novel collaborative knowledge transfer method [23]. They train a compact CNN to remove multiple weather conditions by transferring knowledge from large-scale specific-weather-type neural networks. Zhang et al. propose a universal enhancement network to improve further perception results [56]. Though these methods can achieve encouraging results in several weather types, they are ineffective in the case of a mix of different weather conditions.

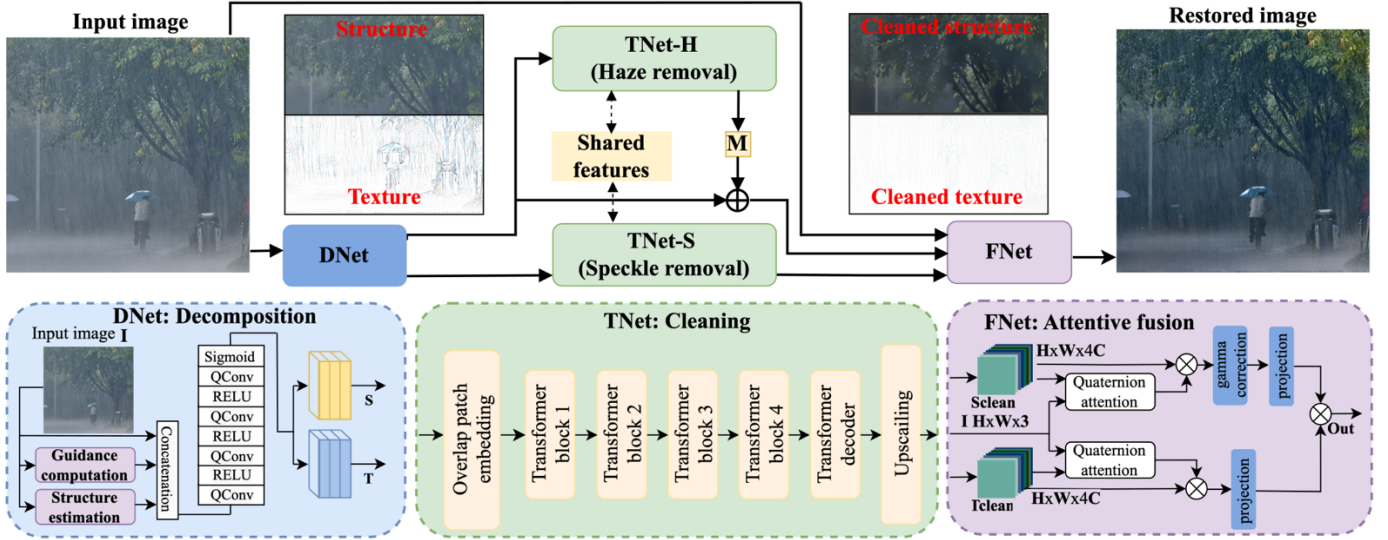

**Fig. 2.** The proposed framework for CMAWRNet. The enhancement process is divided into decomposition, transformation, and fusion. In the decomposition step, a subnetwork DNet decomposes the input image into texture and structure—two separate instances of encoder-decoder based TNet clean texture and structure images in the following cleaning step. Feature sharing between instances of TNet is introduced to ensure adequate processing. Finally, we reconstruct the image with the FNet.

## B. Quaternion Neural Networks

A quaternion number $\hat{q} = a + b\boldsymbol{i} + c\boldsymbol{j} + d\boldsymbol{k}$ extends the concept of the complex number having one real (a) and three imaginary (b, c, d) components, where $a, b, c, d \in \mathbb{R}$ and unity vectors $\boldsymbol{i}, \boldsymbol{j}, \boldsymbol{k}$ ($\boldsymbol{i}^2 = \boldsymbol{j}^2 = \boldsymbol{k}^2 = \boldsymbol{ijk} = -1$) form the quaternion basis. The color input image of the size W by H pixels is represented as a quaternion matrix $\hat{\boldsymbol{I}} \in \mathbb{H}^{H \times W}$:

$$\hat{\boldsymbol{I}} = \mathbf{L} + \mathbf{R}\boldsymbol{i} + \mathbf{G}\boldsymbol{j} + \mathbf{B}\boldsymbol{k} \qquad (1)$$

Here $\mathbf{L}, \mathbf{R}, \mathbf{G}, \mathbf{B} \in \mathbb{R}^{H \times W}$ are real-valued matrices representing luminosity, red, green, and blue channels. Similarly, intermediate feature maps are represented as a group of quaternion-valued matrices.

The quaternion algebra on $\mathbb{H}$ defines operations among quaternion numbers: addition, conjunction, and absolute value, similar to the algebra on complex numbers [30], [31]. The Hamiltonian product defines the non-commutative multiplication of two quaternions $\hat{x} = a_1 + b_1\boldsymbol{i} + c_1\boldsymbol{j} + d_1\boldsymbol{k}$ and $\hat{y} = a_2 + b_2\boldsymbol{i} + c_2\boldsymbol{j} + d_3\boldsymbol{k}$ as:

$$\begin{aligned}
\hat{x} \otimes \hat{y} = &(a_1 a_2 - b_1 b_2 - c_1 c_2 - d_1 d_2) \\
&+ (a_1 b_2 + b_1 a_2 + c_1 d_2 - d_1 c_2)\boldsymbol{i} \\
&+ (a_1 c_2 - b_1 d_2 + c_1 a_2 - d_1 b_2)\boldsymbol{j} \\
&+ (a_1 d_2 + b_1 c_2 - c_1 b_2 - d_1 a_2)\boldsymbol{k}.
\end{aligned} \qquad (2)$$

In QNNs the Hamilton product replaces the real-valued dot product as the transformation between two quaternion-valued feature maps. It ensures the maintenance and exploitation of relations within components of a quaternion feature map.

The convolution of the input $\hat{q} = q_0 + q_1\boldsymbol{i} + q_2\boldsymbol{j} + q_3\boldsymbol{k}$ and kernel $\widehat{W} = W_0 + W_1\boldsymbol{i} + W_2\boldsymbol{j} + W_3\boldsymbol{k}$ is defined as:

$$\hat{q}' = \widehat{W} \otimes \hat{q}. \qquad (3)$$

Typically, the quaternion convolution is implemented as a grouped real-valued convolution. A $C$-channel quaternion feature map is represented as a $4 \cdot C$-channel real-valued feature map. The first $C$ channels represent real components of quaternion feature maps, and the following three groups of $C$ channels each represent $i$, $j$, and $k$-components. The components of weight $\widehat{W}$ is convolved with multiple quaternion inputs.

As can be seen from the examples in Fig. 1, images produced by the reduced by 3.85 parameters quaternion neural network are almost indistinguishable from the ground truth in terms of color. The QNN also preserves the structure and improves details compared to the real-valued network.

## IV. PROPOSED METHOD

### A. Problem Formulation

The CMAWRNet model, illustrated in Fig. 2., consists of three subnetworks: *DNet* for image decomposition, *TNet* for image cleaning, and *FNet* for image reconstruction with gamma correction.

CMAWRNet follows the algorithm 1. First we decompose the input image $\mathbf{I}$ into structure $\boldsymbol{S}$ and texture $\boldsymbol{T}$ components using the DNet subnetwork. We employ a model formulation similar to Retinex problem to find an appropriate structure $\boldsymbol{S}$ by suppressing texture details $\boldsymbol{T}$ in the input image $\boldsymbol{I}$ [68]:

$$\boldsymbol{I} = \boldsymbol{S} \circ \boldsymbol{T} \qquad (4)$$

Where $\boldsymbol{I} \in [0,1]^{W \times H \times 3}$ is the original RGB image, $\boldsymbol{S} \in [0,1]^{W \times H \times 3}$ is the structure component, $T \in [0,1]^{W \times H \times 3}$ is the texture component, with $W$ and $H$ being the width and height of the image, respectively.

The structure $\boldsymbol{S}$, is then processed by TNet to produce a cleaned version $\boldsymbol{S}_{clean}$. Simultaneously, the texture $\mathbf{T}$, representing fine surface details often degraded by rain streaks, raindrops, and snowflakes, is refined using an encoder-decoder network to remove these artifacts, resulting in $\mathbf{T}_{clean}$. After performing gamma correction on $\boldsymbol{S}_{clean}$ to adjust illumination levels, we use attentive quaternion fusion to effectively combine $\boldsymbol{S}_{clean}$ and $\mathbf{T}_{clean}$.

---

**Algorithm 1** CMAWRNet Image Restoration Algorithm

---

**Require:** Input image $\mathbf{I} \in [0,1]^{W \times H \times 3}$

**Ensure:** Restored image $\mathbf{I}_{clean}$

1:  **Initialization:** Set parameters $\gamma_t$, $\gamma_s$, $\gamma$, and patch size $\Omega$

2:  Compute guidance map: $\mathbf{G} \leftarrow |\nabla \mathbf{I}|^{\gamma_t}$ ▷ Equation (5)

3:  Compute structure map: $\mathbf{S}_0 \leftarrow \frac{1}{|\Omega|} \sum_{(i,j) \in \Omega} |\nabla \mathbf{I}|_{i,j}^{\gamma_s}$ ▷ Equation (6)

4:  Compute texture component: $\mathbf{T}_0 \leftarrow \mathbf{I} \oslash \mathbf{S}_0$ ▷ Since $\mathbf{I} = \mathbf{S}_0 \circ \mathbf{T}_0$

   **Refine components using DNet:**

5:  $\mathbf{S} \leftarrow \text{DNet}(\mathbf{S}_0)$ ▷ Refined structure

6:  $\mathbf{T} \leftarrow \text{DNet}(\mathbf{T}_0)$ ▷ Refined texture

   **Clean components:**

7:  Clean structure component: $\mathbf{S}_{clean} \leftarrow \text{TNet-H}(\mathbf{S})$

8:  Clean texture component: $\mathbf{T}_{clean} \leftarrow \text{TNet-S}(\mathbf{T})$

   **Estimate latent variable from structure:**

9:  $\mathbf{M} \leftarrow \text{Conv1x1}(\mathbf{S}_{clean})$

10: Update structure component: $\mathbf{S}_{clean} \leftarrow \mathbf{M} \circ \mathbf{S}_{clean}$

11: Apply gamma correction: $\mathbf{S}_{clean} \leftarrow \mathbf{S}_{clean}^{\gamma}$

   **Compute attention maps using FNet:**

12: $\mathbf{M}_S \leftarrow \sigma\left(\mathbf{W}_2^{(S)} \text{ReLU}(\mathbf{W}_1^{(S)}[\mathbf{S}_{clean}, \mathbf{I}])\right)$

13: $\mathbf{M}_T \leftarrow \sigma\left(\mathbf{W}_2^{(T)} \text{ReLU}(\mathbf{W}_1^{(T)}[\mathbf{T}_{clean}, \mathbf{I}])\right)$

   **Fuse components:**

14: Fuse structure: $\mathbf{S}_{fused} \leftarrow \mathbf{M}_S \otimes \mathbf{S}_{clean}$

15: Fuse texture: $\mathbf{T}_{fused} \leftarrow \mathbf{M}_T \otimes \mathbf{T}_{clean}$

16: Compute restored image: $\mathbf{I}_{clean} \leftarrow \mathbf{S}_{fused} \circ \mathbf{T}_{fused}$

17: **return** $\mathbf{I}_{clean}$

---

The clean image is restored as $\mathbf{I}_{clean} = \mathbf{S}_{clean} \circ \mathbf{T}_{clean}$. CMAWRNet offers an effective solution to the ill-posed problem of restoring images affected by adverse weather conditions by processing the decomposed components individually and integrating them through attentive fusion.

### B. DNet: Structure and Texture Decomposition

In this subsection, we introduce DNet, which decomposes an image $\mathbf{I}$ into structure $\mathbf{S}$ and texture $\mathbf{T}$ components. Following the Retinex model concept [63], we use exponentiated local derivatives to compute the guidance and structure maps. The guidance map is computed as:

$$\mathbf{G} = |\nabla \mathbf{I}|^{\gamma_t} \quad (5)$$

where $|\nabla \mathbf{I}|$ represents the magnitude of the local gradient of the image, and $\gamma_t$ adjusts the influence of the gradient magnitude on $\mathbf{G}$, affecting its sensitivity to texture details. We estimate the initial structure map using the local average of the exponentiated gradient magnitude:

$$\mathbf{S}_0 = \frac{1}{|\Omega|} \sum_{(i,j) \in \Omega} |\nabla \mathbf{I}|_{i,j}^{\gamma_s} \quad (6)$$

where $\Omega$ is a local patch of size $3 \times 3$ around each pixel of $\mathbf{I}$, and $\gamma_s$ modifies the impact of the gradient magnitude in $\mathbf{S}_0$, emphasizing structural components.

In our experiments, we use $\gamma_t = 0.5$ and $\gamma_s = 1.5$ to moderately enhance fine details without overly exaggerating texture components. The initial texture component is then computed as $\mathbf{T}_0 = \mathbf{I} \oslash \mathbf{S}_0$ since $\mathbf{I} = \mathbf{S}_0 \circ \mathbf{T}_0$.

Finally, refining the decomposition, we apply a quaternion convolutional refinement network DNet to both $\mathbf{S}_0$ and $\mathbf{T}_0$. This network has three convolutional layers with ReLU activation functions, followed by a sigmoid activation function in the final layer to ensure the output range is between 0 and 1.

### C. TNet: Lightweight Quaternion Encoder-Decoder

This subsection introduces TNet, a lightweight quaternion encoder-decoder network. First, we perform overlapping patch embedding on the input image of size $H \times W \times 3$ [64]. Then, four quaternion transformers are used to obtain a hierarchical feature map $\mathbf{F}$. We employ multi-head self-attention layers and quaternion feed-forward networks in each transformer block to calculate self-attention features:

$$T_i(\mathbf{I}_i) = \text{QFFN}(\text{QMSA}(\mathbf{I}_i) + \mathbf{I}_i) \quad (7)$$

where $T_i$ represents the transformer block at stage i, QFFN denotes the quaternion feed-forward network block, QMSA stands for quaternion multi-head self-attention, $\mathbf{I}_i$ is the input at stage i in the encoder [65].

The queries $Q$, keys $K$, and values $V$ are generated by the quaternion feed-forward network, and the attention is:

$$\text{QMSA}(Q, K, V) = \text{softmax}\left(\frac{Q \otimes K^T}{\sqrt{d}}\right) \otimes V \quad (8)$$

where d represents the dimensionality, $\otimes$ denotes quaternion multiplication. The computation within the quaternion feed-forward network block is summarized as:

$$\text{QFFN}_i(X_i) = \text{QMLP}\left(\text{GELU}\left(\text{DWC}\left(\text{QMLP}(X_i)\right)\right)\right) + X_i \quad (9)$$

where $X_i$ refers to the self-attention features at stage i, DWC is depth-wise convolution [66], GELU is the Gaussian Error Linear Unit activation function [67], QMLP is a quaternion multi-layer perceptron.

Features extracted by the fourth transformer for both TNet-H and TNet-S are concatenated and used as input for decoders in both instances. TNet-H and TNet-S are identical, except that TNet-H is coupled with a single $\text{Conv}_{1 \times 1}$ layer to estimate the latent variable $M$ [4] from the equation:

$$J(x) = \mathbf{M}(x) \circ \mathbf{I}(x) \quad (10)$$

Where $J$ is the restored image, $\mathbf{I}$ is the input image, $x$ represents spatial coordinates, $\mathbf{M}(x)$ is a learnable latent variable dependent on $\mathbf{I}$ [4]. The latent variable $M(x)$ is estimated as follows:

$$\mathbf{M}(x) = \frac{S(x) + tA - A}{tI(x)} \quad (11)$$

where $A$ is the airlight, $t$ is the transmission map, $\mathbf{S}(x)$ represents the structure component.

TNet-H generates the clean version of the structure components, while TNet-S generates the clean version of the texture components. These outputs are further fused by FNet, as shown in Fig. 2.

### D. FNet: Attentive fusion

The images decomposed into structure $\mathbf{S}$ and texture $\mathbf{T}$ components need to be recombined. The cleaned features $\mathbf{F}_S$ and $\mathbf{F}_T$, along with the encoded features of the damaged image $\mathbf{I}$, are fed into the attentive fusion subnetwork, FNet.

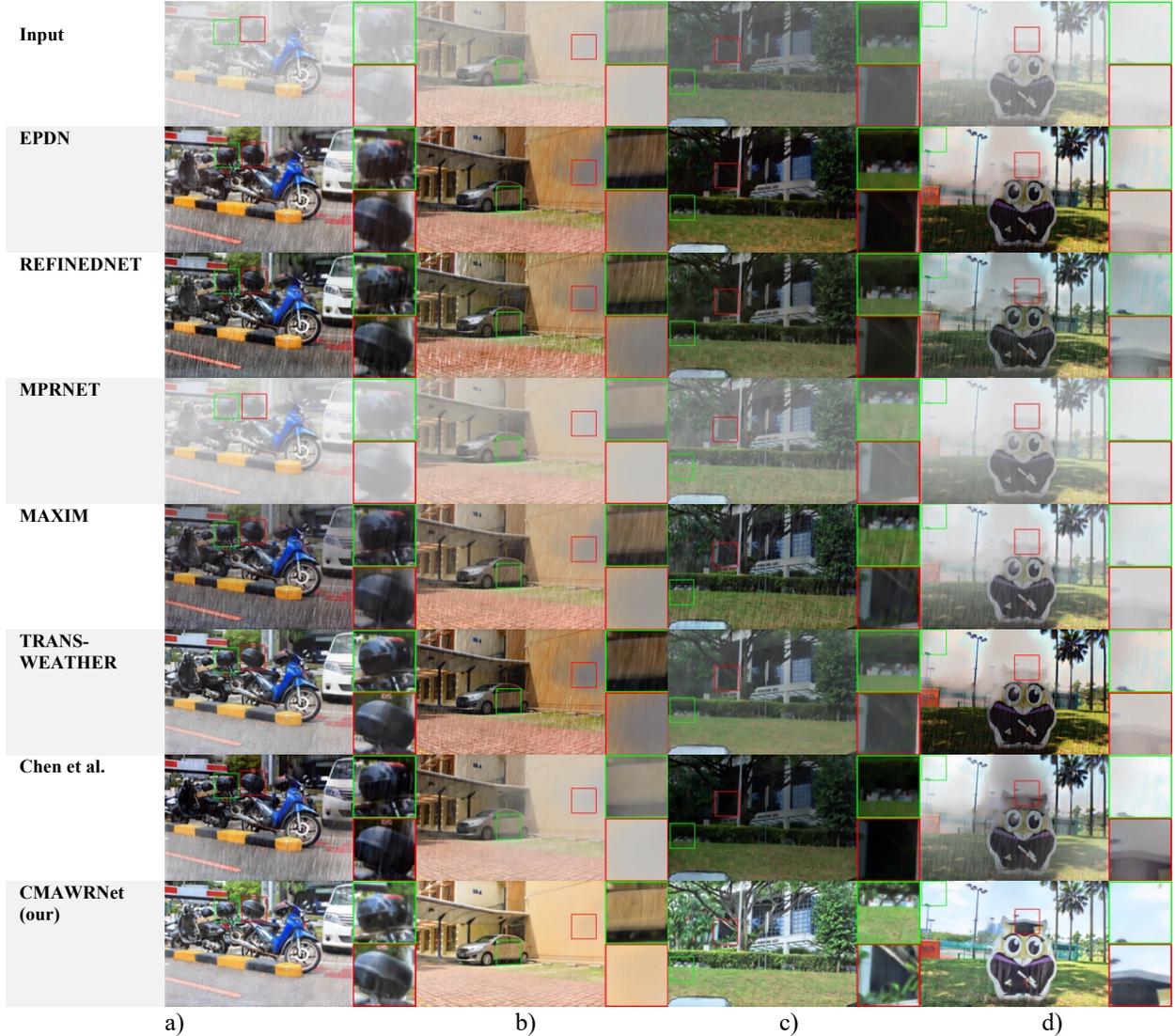

**Fig. 3.** Comparison of synthetic rain and haze removal by CMAWRNet and state-of-the-art methods. Specialized haze removal techniques, EPDN and RefineDNet, effectively remove haze from most images but leave rain streaks and produce darker images with loss of detail in dark regions, especially struggling with non-homogeneous haze structures like in image d). MPRNet removes rain streaks but yields low-contrast images with residual fog, making them less visually appealing. MAXIM is more effective at fog removal but struggles with removing rain streaks. CMAWRNet effectively reduces both fog and rain degradations, producing vivid images with preserved details, particularly in dark regions, and outperforms other methods in terms of mAP/mA

FNet uses attention masks to produce importance weights for the fusion process. The fused feature map $F_o$ is computed as:

$$\boldsymbol{F_o} = \boldsymbol{M_S} \otimes \boldsymbol{F_S} + \boldsymbol{M_T} \otimes \boldsymbol{F_T} \tag{12}$$

Where $\boldsymbol{M_S}$ and $\boldsymbol{M_T}$ are attention maps for the structure and texture components, respectively, $\otimes$ denotes element-wise multiplication.

The attention maps $\boldsymbol{M_S}$ and $\boldsymbol{M_T}$ for quaternion inputs are computed as:

$$\boldsymbol{M}(f) = \text{Sigmoid}\big(W_2 \,\text{ReLU}(W_1 f)\big) \tag{13}$$

Here $W_1$ and $W_2$ are trainable weights, $f$ represents the concatenated features of the input components.

The combined feature map $\boldsymbol{F_o}$ is further fed into a $\text{Conv}_{3\times3}$ layer to project it to a single output quaternion image. This output is then concatenated with the input feature map and the damaged image $\boldsymbol{I}$ to produce the final restored image.

### E. QSSIM Loss Function

SSIM-loss was found to be used in deep learning, but SSIM ignores color information. Instead, we use the quaternion version of QSSIM, defined as [57]:

$$QSSIM_{gt,rec} = \left(\frac{2\mu_{q_{gt}} \cdot \mu_{q_{rec}}}{\mu_{q_{gt}}^2 + \mu_{q_{rec}}^2}\right)\left(\frac{\sigma_{q_{gt,rec}}}{\sigma_{q_{gt}}^2 + \sigma_{q_{rec}}^2}\right) \tag{14}$$

Here "gt" means ground truth image, and "rec" – reconstructed image. Quaternion means $\mu_{q_{ref}}$, $\mu_{q_{rec}}$, variations $\sigma_{q_{gt}}$, $\sigma_{q_{rec}}$ and covariance $\sigma_{q_{gt,rec}}$ are computed as in [57].

## IV. EXPERIMENTAL RESULTS

In this section, we evaluate our method on several image restoration tasks, including snow removal, rain-streak removal,

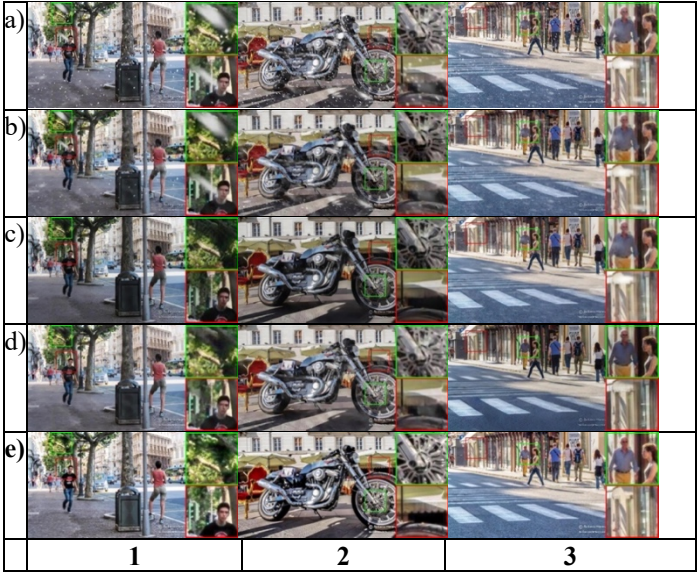

**Fig. 4.** Synthetic snow images: a) – input image, b) – DesnowNet, c) TransWeather, d) – CMAWRNet, e) Ground truth. In the image (1) all the state of art methods remove small snow particles, but DesnowNet struggles to remove the large ones. CMAWRNet produces realistic images in place of large snow particles. Simulation results on the image (2) show that our CMAWRNet network removes all the snow particles and preserves small details in the wheel area compared with the state of art methods. Also, in the image (3) only CMAWRNet successfully removes snow particles from the road, producing an image close to ground truth.

and adherent raindrop removal on a large publicly available benchmarking synthetic and real-world dataset, and compare our performance to state-of-the-art algorithms.

### A. Dataset and Training

The CMAWRNet is implemented in PyTorch [59] and trained in two steps on a single Nvidia Tesla A100 GPU. First, the *DNet* and *FNet* are trained to perform texture/structure split and low-light correction on the LOL (Low-Light) dataset [58]. The LOL dataset consists of 500 image pairs of low-light and normal-light scenes, primarily indoors, with a resolution of 400×600 pixels. It is divided into 485 training pairs and 15 testing pairs. The model is trained for 100 epochs, starting with a learning rate 0.001, halved every 25 epochs.

In the second stage, we train the entire network using a combination of bad weather datasets. The "RainDrop" dataset contains 1,119 pairs of clean images and images with adherent raindrops [9]. The "Snow100K" dataset provides 50,000 training images and 50,000 validation images with synthetic snow [13]. The "Outdoor-Rain" dataset consists of 9,000 training samples and 1,500 validation samples, combining synthetic rain streaks with haze. Since the datasets vary in size, we sample 9,000 random images from "Snow100K" and oversample the "RainDrop" dataset by applying random data augmentation, including rotation, random cropping, and affine distortion. During the first 50 epochs, the weights of DNet and FNet are frozen. The initial learning rate is set to $1 \times 10^{-3}$ and is gradually reduced to $1 \times 10^{-7}$ using a cosine annealing strategy over a total of 200 epochs. During training, we randomly sample 256×256 pixel patches from the original-resolution images. At inference time, our model process images of varying sizes by dividing them into overlapping blocks.

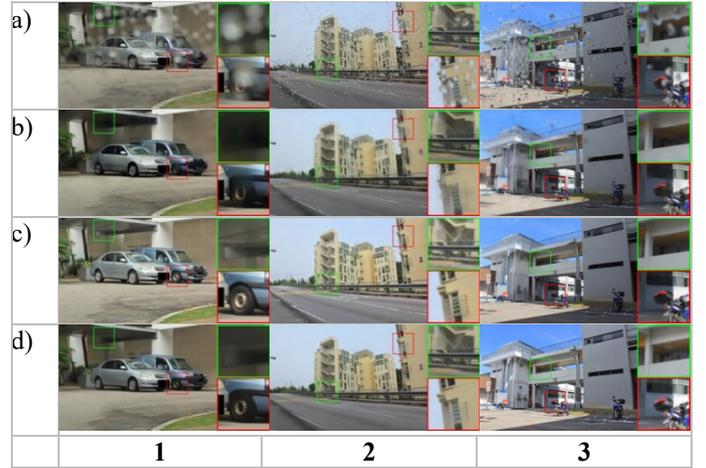

**Fig. 5.** Rain drop images. a) Input, b) – TransWeather, c) CMAWRNet, d) Ground Truth. Both TransWeather and CMAWRNet successfully remove raindrops, but CMAWRNet produces a brighter image with better details.

TABLE I
QUANTITATIVE COMPARISON ON THE TEST1 (RAIN+FOG) DATASET BASED ON PSNR AND SSIM.

| Type | Method | PSNR | SSIM |
|---|---|---|---|
| Task specific | EPDN [59] | 13.36 | 0.5830 |
| | RefineDNet [5] | 15.68 | 0.6400 |
| Multi-degradation | MPRNet [16] | 21.90 | 0.8456 |
| | MAXIM [60] | 26.91 | 0.9212 |
| Universal | TransWeather [22] | 27.96 | 0.9509 |
| | Chen et al. [23] | 28.18 | 0.9524 |
| | CMAWRNet (Our) | **30.02** | **0.9654** |

TABLE II
QUANTITATIVE COMPARISON ON THE SNOWTEST100K-L TEST DATASET BASED ON PSNR AND SSIM

| Type | Method | PSNR | SSIM |
|---|---|---|---|
| Task-specific | JSTASR [54] | 25.32 | 0.8076 |
| | DesnowNet [13] | 27.17 | 0.8983 |
| Universal | Chen et al. [23] | 28.33 | 0.8820 |
| | TransWeather [22] | 28.48 | 0.9308 |
| | CMAWRNet (Our) | **30.08** | **0.9458** |

TABLE III
QUANTITATIVE COMPARISON ON THE RAINDROP TEST DATASET BASED ON PSNR AND SSIM

| Type | Method | PSNR | SSIM |
|---|---|---|---|
| Task-specific | Attn. GAN [9] | 30.55 | 0.9023 |
| | Quan et al. [53] | 31.33 | 0.9263 |
| Universal | Chen et al.[23] | 28.84 | 0.9460 |
| | Transweather [22] | 31.12 | 0.9268 |
| | CMAWRNet | **32.43** | **0.9518** |

### B. Comparison with state of the art

We compare CMAWRNet to state-of-the-art methods on synthetic and real-world image datasets. The quantitative results are evaluated with PSNR and SSIM [46].

TABLE IV - AVERAGE MEAN PRECISION AND RECALL ON SUBSETS OF DAWN DATASET

| | DAWN Fog | | | | DAWN Rain | | | | DAWN Snow | | | |
|---|---|---|---|---|---|---|---|---|---|---|---|---|
| | mAP | mAPs | mAR | mARs | mAP | mAPs | mAR | mARs | mAP | mAPs | mAR | mARs |
| Baseline | 0.548 | 0.076 | 0.422 | 0.183 | 0.520 | 0.115 | 0.467 | 0.221 | 0.593 | 0.124 | 0.436 | 0.212 |
| EPDN | 0.551 | 0.082 | 0.426 | 0.187 | 0.523 | 0.118 | 0.472 | 0.228 | 0.598 | 0.127 | 0.442 | 0.216 |
| RefineDNet | 0.549 | 0.076 | 0.423 | 0.183 | 0.520 | 0.115 | 0.468 | 0.222 | 0.593 | 0.124 | 0.436 | 0.212 |
| MPRNet | 0.539 | 0.054 | 0.405 | 0.168 | 0.522 | 0.116 | 0.471 | 0.224 | 0.583 | 0.117 | 0.421 | 0.196 |
| Maxim | 0.550 | 0.080 | 0.427 | 0.189 | 0.518 | 0.113 | 0.464 | 0.216 | 0.594 | 0.127 | 0.424 | 0.215 |
| TransWeather | 0.552 | 0.081 | 0.441 | 0.181 | 0.523 | 0.168 | 0.474 | 0.236 | 0.572 | 0.109 | 0.432 | 0.198 |
| Chen et. al. | 0.488 | 0.026 | 0.347 | 0.058 | 0.491 | 0.039 | 0.406 | 0.232 | 0.572 | 0.109 | 0.432 | 0.198 |
| JSTASR | 0.547 | 0.076 | 0.422 | 0.183 | 0.520 | 0.115 | 0.467 | 0.221 | 0.598 | 0.128 | 0.439 | 0.217 |
| DesnowNet | 0.548 | 0.075 | 0.421 | 0.179 | 0.520 | 0.115 | 0.467 | 0.220 | 0.597 | 0.125 | 0.438 | 0.216 |
| Attn Gan | 0.548 | 0.076 | 0.422 | 0.183 | 0.520 | 0.115 | 0.467 | 0.221 | 0.593 | 0.124 | 0.436 | 0.212 |
| Quan et al. | 0.548 | 0.076 | 0.422 | 0.183 | 0.520 | 0.115 | 0.467 | 0.221 | 0.593 | 0.124 | 0.436 | 0.212 |
| QSAM-Net [69] | 0.549 | 0.076 | 0.431 | 0.184 | 0.531 | 0.114 | 0.475 | 0.225 | 0.593 | 0.124 | 0.436 | 0.212 |
| LQC [70] | 0.578 | 0.090 | 0.451 | 0.194 | 0.531 | 0.181 | 0.478 | 0.234 | 0.598 | 0.155 | 0.445 | 0.215 |
| CMAWRNet | 0.587 | 0.132 | 0.471 | 0.221 | 0.536 | 0.198 | 0.482 | 0.249 | 0.614 | 0.164 | 0.451 | 0.218 |

**Rain and Fog:** We evaluate the method on the synthetic dataset Test1 (part of Outdoor-Rain) [57]. We compare with baseline methods for dehazing: EPDN [59], RefineDNet [5]; multi-degradation rain-removal: MPRNET [16], MAXIM [60] and universal: TransWeather [22], Chen et al. [23]. The produced results are shown in Fig. 3. One can see that CMAWRNet is effective for a combination of haze and rain streaks and produces more vivid images with better visibility, especially in darker regions. Other methods struggle with this combination. Although Chen et al. and Transweather are suitable for various types of weather, they cannot process the combination of different weather types. The quantitative results are presented in Table I. CMAWRNet outperforms other methods by a large margin.

**Snow:** The results of various methods on synthetic images from the Snow100K dataset [13] are shown in Fig. 4. The quantitative results are presented in Table II. The visual analysis shows that the result produced by CMAWRNet contains less residual haze. Our method can remove more snow particles and rain streaks than other methods. RefinedDNet effectively removes the fog but produces dark images. Also, it cannot remove snow particles in the image (1). Chen et al. remove most snow particles but fail to remove the fog, producing dark artifacts. Transweather removes the snow particles but does not improve visibility at all.

**RainDrop Images.** Fig. 5 presents the simulation results of various methods tested on the RainDrop dataset [9]. The quantitative results are presented in Table IV.

**Object detection:** The presence of adverse weather conditions influences the functioning of downstream applications, such as object detection. To evaluate the influence of weather removal on object detection, we compare the performance of SCNet [65] object detector on the DAWN dataset [66]. As the SCNet was trained on the COCO dataset for metrics computations, we only consider object categories from DAWN. The DAWN dataset contains 250 real-world traffic images for each weather condition: fog, snow, and rain. Each image is annotated with bounding boxes of cars, trucks, autobuses, motorcycles, and pedestrians. The quantitative comparison is presented in Table III. Mean average precision mAP and mAR are computed for the intersection of union IOU of 50%[67].

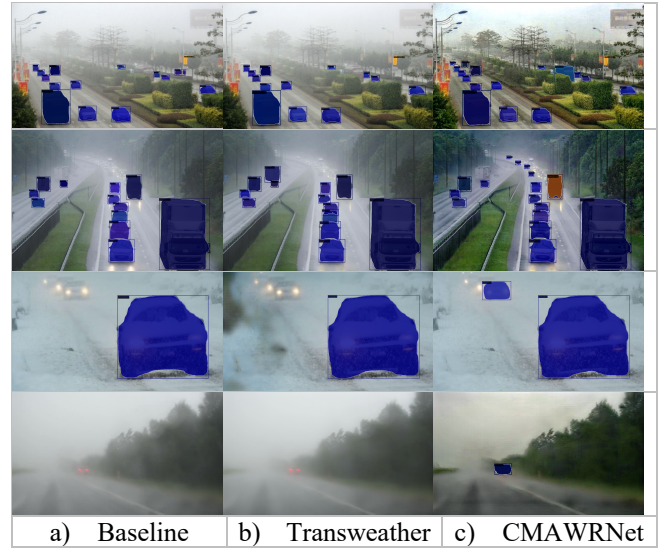

a) Baseline   b) Transweather   c) CMAWRNet

**Fig. 6.** Detection results on the image without preprocessing (Baseline), Transweather, and CMAWRNet. As can be seen, CMAWRNet improves the detection of small background objects in fog, snow, and rain conditions.

As can be seen, CMAWRNet improves the detection performance in all weather conditions, especially the detection of small objects, as shown by the values of mAPs and mARs. A comparison of object detection on various real-world images is presented in Fig. 7. The visual analysis shows that the result produced by CMAWRNet contains less residual haze. Also, our method can remove more snow particles and rain streaks than other methods. RefinedDNet effectively removes the fog but produces dark images. Also, it cannot remove snow particles in the image (1). Chen et al. remove most of the snow particles but fail to remove the fog, producing dark artifacts. Transweather removes the snow particles but does not improve visibility at all.

**Real Images:** In Fig. 7, we present the visual results recovered by the proposed method under haze, snow, and rain scenarios compared to state-of-the-art methods. Our method achieves remarkable visual quality for various types of weather.

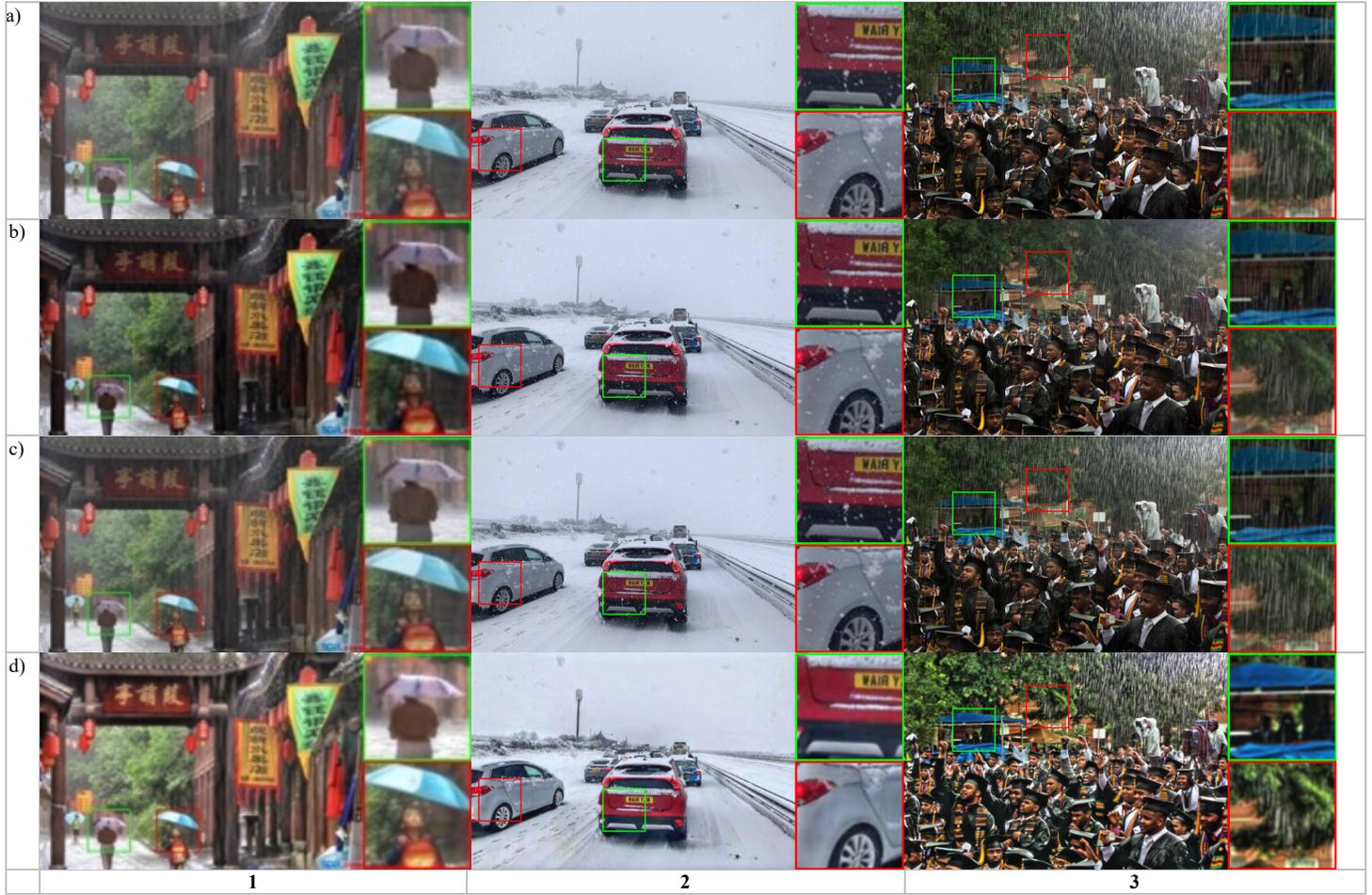

**Fig. 7.** a) input image, b) Chen et. al. c) TransWeather, d) CMAWRNet. CMAWRNet generally produces vivid images with better visibility. CMAWRNet brings up better background details in the image (1). Moreover, in image (2) both TransWeather and Chen et al. methods failed to remove the snow particles, while CMAWRNet succeeded. CMARNet removes more rain streaks in the image (3) compared to other methods

TABLE V - AVERAGE MEAN PRECISION AND RECALL ON SUBSETS OF DAWN DATASET

| Configuration | DNet | Shared Features | Quaternion Enhancement | QSSIM Loss Function | Params (M) | PSNR | SSIM | Inference Time (ms) |
|---|---|---|---|---|---|---|---|---|
| Standalone Transformer (TNet only) | ✗ | ✗ | ✗ | ✗ | 5.0 | 27.10 | 0.9320 | 30.5 |
| No Image Decomposition | ✗ | ✓ | ✓ | ✓ | 13.5 | 28.30 | 0.9440 | 28.7 |
| No Shared Features | ✓ | ✗ | ✓ | ✓ | 14.8 | 29.05 | 0.9532 | 26.8 |
| Real-Network | ✓ | ✓ | ✗ | ✓ | 16.2 | 29.62 | 0.9519 | 25.6 |
| No QSSIM Loss Function | ✓ | ✓ | ✓ | ✗ | 18.0 | 29.85 | 0.9580 | 24.7 |
| **CMAWRNet** | ✓ | ✓ | ✓ | ✓ | 18.0 | 30.02 | 0.9654 | 24.3 |

## C. Ablation analysis

We conducted an ablation study on the "Outdoor-Rain" dataset [57] to evaluate the effectiveness of each component in our CMAWRNet architecture. Table V summarizes the results, including average inference times and number of parameters for different co, illustrating the trade-offs between accuracy and processing speed associated with different architectural choices.

As shown in Table V, each component of the CMAWRNet architecture contributes significantly to overall performance. The Standalone Transformer (TNet only) serves as a baseline, emphasizing the necessity of integrated image decomposition and feature sharing for optimal results. Excluding the Image Decomposition Network (No DNet) demonstrates that the network relies on image decomposition for effective operation, as its absence leads to reduced performance. The configuration without Shared Features indicates that while individual components are beneficial, their collective operation with shared features substantially enhances results. Removing Quaternion Enhancements (Real-Network) highlights the integral role of quaternion algebra in managing complex image characteristics and improving quality.

Excluding the QSSIM loss function shows its importance in accurate image quality assessment and enhancement. The full architecture, which includes all components, achieves the best performance with the lowest inference time, confirming the effectiveness of integrating all developed modules.

### D. Complexity analysis

The evaluation was performed on a single NVIDIA A100 GPU. The methods compared include Chen et al. [23], TransWeather [22], and Lightweight Quaternion Chebyshev, in addition to our own.

TABLE VI
COMPLEXITY ANALYSIS

| Methods | Average Inference Time (ms) |
|---|---|
| Chen et al. [23] | 45.2 |
| TransWeather [22] | 37.8 |
| Lightweight Quaternion Chebyshev | 29.5 |
| CMAWRNet (Ours) | 24.3 |

The table VI presents the average inference time measured in milliseconds (ms) for processing a single image of a standard resolution (1920x1080). CMAWRNet, achieves a lower inference time than the other evaluated methods, suggesting it is more efficient for real-time applications or scenarios where computational resources are limited. This efficiency gain does not compromise the quality of weather removal, as demonstrated in our earlier qualitative and quantitative evaluations. By integrating advanced techniques and optimizations specific to the architecture of the NVIDIA A100 GPU, we've managed to reduce the computational burden while maintaining high performance in weather condition removal tasks.

## V. CONCLUSION

We proposed CMAWRNet – an efficient, multistage architecture for adverse weather removal in this work. We build a single architecture focused on processing multiple weather conditions by decomposing the input image into texture and structure components and processing them separately. To this end, we propose a novel quaternion neural network CMAWRNet, DNet – quaternion network for image decomposition, and FNet – quaternion attention-based network for low-light correction. CMAWRNet delivers significant performance gains on various benchmark datasets. We also obtain better visual results on real-world images of snow and rain.

## ACKNOWLEDGMENT

The work was supported by the U.S. Department of transportation, federal highway administration (FHWA), under contract 693jj320c000023.

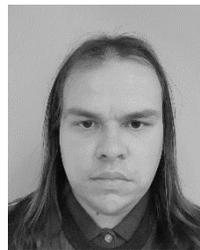

**Vladimir Frants** received a bachelor's degree (B.E.) in electrical engineering from the South Russian University of Economics and service, Russia, in 2011 and an M.S. degree in electrical engineering from the Don State Technical University, Russia, in 2013. He is pursuing a Ph.D. in computer science with The Graduate Center CUNY, USA. His current research interests include artificial intelligence, computer vision, image processing, and machine learning.

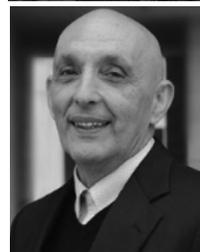

**Sos S. Agaian**, (Fellow, IEEE) is currently a Distinguished Professor with The City University of New York/CSI. His research interests include computational vision, machine learning, multimedia security, multimedia analytics, biologically inspired image processing, multi-modal biometrics, information processing, image quality, and biomedical imaging. He has authored over 750 technical articles and ten books in these areas. He is also listed as a co-inventor on 54 patents/disclosures. The technologies he invented have been adopted by multiple institutions, including the U.S. Government, and commercialized by industry. He is a fellow of SPIE, IS&T, AAIA, and AAAS. He received the Maestro Educator of the year, sponsored by the Society of Mexican American Engineers. He received the Distinguished Research Award at The University of Texas at San Antonio. He was a recipient of the Innovator of the Year Award, in 2014, the Tech Flash Titans-Top Researcher Award (San Antonio Business Journal), in 2014, the Entrepreneurship Award (UTSA-2013 and 2016), and the Excellence in Teaching Award, in 2015. He is an Editorial Board Member for the Journal of Pattern Recognition and Image Analysis and an Associate Editor for several journals, including the IEEE

Transactions On Image Processing, The IEEE Transactions On Systems, Man And Cybernetics, Journal of Electrical and Computer Engineering (Hindawi Publishing Corporation), International Journal of Digital Multimedia Broadcasting (Hindawi Publishing Corporation), and Journal of Electronic Imaging (IS&T and SPIE).

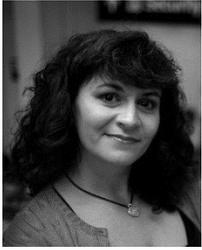 **Karen Panetta** (Fellow, IEEE) received a B.S. degree in computer engineering from Boston University, Boston, MA, USA, and an M.S. and Ph.D. degrees in electrical engineering from Northeastern University, Boston. She is currently the Dean of graduate engineering education and a Professor with the Department of Electrical and Computer Engineering. with secondary appointments in computer science and mechanical engineering at Tufts University, Medford, MA, USA, and the Director of the Dr. Panetta's Vision and Sensing System Laboratory. Her research interests include developing efficient algorithms for simulation, modeling, signal, and image processing for biomedical and security applications. She was a recipient of the 2012 IEEE Ethical Practices Award and the Harriet B. Rigas Award for Outstanding Educator. In 2011, she was awarded the Presidential Award for Engineering and Science Education and Mentoring by U.S. President Obama. She is the Vice President of SMC, Membership, and Student Activities. She was the President of the 2019 IEEE-HKN. She is the Editor-in-Chief of the IEEE Women in Engineering Magazine. She was the IEEE-USA Vice-President of communications and public affairs. She is a Fellow of the National Academy of Inventors, AAIA, AAAS and NASA JOVE. She is a member of the European Academy of Sciences and Arts.